%% file: main.tex
\documentclass[11pt,letterpaper]{article}

\usepackage[letterpaper,margin=1in]{geometry}

\usepackage{mathptmx}          
\usepackage[T1]{fontenc}
\usepackage[utf8]{inputenc}
\usepackage{textcomp}          
\usepackage{lmodern}           

\usepackage{setspace}
\makeatletter
\let\origtextunderscore\_
\renewcommand{\_}{\origtextunderscore\allowbreak}
\makeatother
\usepackage{graphicx}
\usepackage{booktabs}
\usepackage{array}
\usepackage{caption}
\usepackage{enumitem}
\usepackage{xcolor}
\usepackage{tikz}
\usetikzlibrary{shapes.geometric, arrows.meta, positioning, fit}
\usepackage{hyperref}
\hypersetup{
    colorlinks=true,
    linkcolor=black,
    urlcolor=black,
    citecolor=black,
    pdftitle={A Reliability Assessment of LALM Audio Judges for Full-Duplex Voice Agents},
    pdfauthor={A. Sayyad, J. Emmons, S. Jones, T. Lin, H. Krishnan}
}

\usepackage{titlesec}
\titleformat{\section}
    {\normalfont\fontsize{13}{15}\bfseries}
    {\thesection}{1em}{}
\titleformat{\subsection}
    {\normalfont\fontsize{11}{13}\bfseries\itshape}
    {\thesubsection}{1em}{}
\titlespacing*{\section}{0pt}{14pt}{4pt}
\titlespacing*{\subsection}{0pt}{8pt}{2pt}

\makeatletter
\newcommand{\papertitle}[1]{%
    \begin{center}%
    {\fontsize{18}{22}\bfseries #1\par}%
    \end{center}%
    \vspace{2pt}%
}
\newcommand{\paperauthors}[2]{%
    \begin{center}%
    {\fontsize{12}{14}\selectfont #1\par}%
    \vspace{2pt}%
    {\fontsize{11}{13}\itshape #2\par}%
    \end{center}%
    \vspace{6pt}%
}
\newenvironment{paperabstract}{%
    \begin{center}\fontsize{11}{13}\bfseries Abstract\par\end{center}%
    \vspace{2pt}%
    \begingroup
    \leftskip=0.5in \rightskip=0.5in
    \fontsize{10}{12}\selectfont
    \setstretch{1.15}%
    \setlength{\parindent}{0pt}%
    \noindent\ignorespaces
}{%
    \par\endgroup
    \vspace{4pt}%
}
\newcommand{\paperkeywords}[1]{%
    \begingroup
    \leftskip=0.5in \rightskip=0.5in
    \fontsize{10}{12}\selectfont
    \setlength{\parindent}{0pt}%
    \noindent{\bfseries Keywords:} #1\par
    \endgroup
    \vspace{10pt}%
}
\makeatother

\newenvironment{references}{%
    \begingroup
    \fontsize{10}{12}\selectfont
    \setstretch{1.03}%
    \setlength{\parindent}{0pt}%
    \begin{list}{}{%
        \setlength{\leftmargin}{0.30in}%
        \setlength{\itemindent}{-0.30in}%
        \setlength{\itemsep}{2pt}%
        \setlength{\parsep}{0pt}%
    }%
}{%
    \end{list}
    \endgroup
}
\newcommand{\refentry}[2]{\item[]\makebox[0.30in][l]{[#1]}#2}

\begin{document}

\thispagestyle{empty}

\papertitle{A Reliability Assessment of LALM Audio Judges
for Full-Duplex Voice Agents}

\paperauthors{A.~Sayyad, J.~Emmons, S.~Jones, T.~Lin, H.~Krishnan}{Salesforce Applied AI Research, eVerse team}

\begin{paperabstract}
We report the empirical reliability of Gemini models as audio judges
that score full-duplex agent conversations directly from the raw
stereo waveform, tested across three models in the Gemini family: 2.5
Flash, 3.5 Flash, and 3.1 Pro. Our primary evidence base uses Gemini
2.5 Flash as the ground-truth model, validated against three
calibrated human raters on 209 stereo sessions, scored on 8 production
dimensions: 152 full-duplex conversations across 13 accent-and-condition
strata, together with 57 adversarial defect-injected clips. The
evidence for Gemini 2.5 Flash is consistent across three tests. (i) On
5 of 8 dimensions the LALM-human Spearman $\rho$ departs from the
pairwise human-human $\rho$ by at most 0.07, and on 7 of 8 dimensions
the two quantities' 95 percent bootstrap confidence intervals overlap.
(ii) The LALM agrees with the three-rater human mean within 1 point on
60 to 92 percent of sessions on 6 of 8 dimensions. (iii) On 45 of 48
(defect, dimension) cells the LALM is as sensitive as humans or better
under Newcombe-Wilson 95 percent confidence intervals, though most of
these are underpowered nulls rather than demonstrated parity.
Rank-ordering ability transfers across the Gemini family: 3.5 Flash
improves simple agreement to 8 of 8 dimensions, while 3.1 Pro rates
several dimensions markedly lower than humans despite comparable rank
correlation. A model swap should be re-validated on
calibration specifically, not assumed from rank-correlation alone. We
identify four areas where deployment requires care, and we estimate
that human rating alone for our current evaluation cadence costs
roughly two orders of magnitude more than the equivalent LALM
workload. The data presented here provides a defensible empirical
basis for deploying the LALM as a substitute or fourth rater on the
dimensions where the evidence supports it.
\end{paperabstract}

\paperkeywords{LALM-as-judge, audio language models, validation, voice
agents, production deployment, full-duplex audio.}

\section{Background}
Our voice-agent product surface relies on two production audio
judges, AgentSpeechFidelity and ConversationalAudioQuality, which
between them emit ratings on eight dimensions for every evaluated
session. To date these judges have been validated informally against
single raters on small samples. This study addresses a single question:
whether LALM-generated ratings agree with human ratings closely enough
to substitute for one or more human raters in a production
audio-evaluation pipeline. We frame substitutability as an empirical,
per-dimension property rather than a global one, and quantify it against
a three-rater human reference standard.

Recent work in the LALM-as-judge literature reports strong correlations
between LALM judges and human Mean Opinion Score (MOS) on isolated
text-to-speech utterances [1, 2, 3]. To our knowledge, this is the first
study to evaluate a LALM as a judge of raw audio from enterprise
full-duplex agent conversations against a multi-rater human reference
standard. We do so over stereo agent-client conversations where channel
content, turn-taking dynamics, and conversational disfluencies all
influence the rating distribution. We extend this line
of work to full-duplex conversational audio using a production
evaluation stack.

\section{Related Work}
\textit{LALM-as-judge for speech.} AudioJudge [1] judges speech
attributes (pronunciation, rate, quality) with high system-level human
correlation; ALLD [2] produces descriptive MOS judgements; and
SpeechQualityLLM [3] predicts dimension-wise MOS. Broad benchmarks such
as AIR-Bench [10] score free-form outputs with a text LLM rather than
validating an audio judge against humans. All operate on isolated
utterances, and none reports multi-rater human validation over
full-duplex conversational audio.

\textit{Full-duplex evaluation.} Full-Duplex-Bench [8] and its
multi-turn successor [9] score turn-taking, overlap, and latency of
spoken-dialogue systems; the latter judges an automated examiner's
interactions with an LLM reading \emph{transcripts}, not the waveform.
Both target conversational \emph{behaviour} via programmatic or
transcript-based metrics, not per-dimension audio-\emph{fidelity}
ratings judged from the audio against humans. We fill that gap: we
measure how closely a production, audio-native LALM judge's
per-dimension ratings track three calibrated humans on stereo
full-duplex agent-client conversations.

\textit{Reliability versus validity.} We invoke a long-standing
psychometric fact: high raw agreement can coexist with near-zero
chance-corrected agreement when ratings saturate near one end of the
scale, the ceiling degeneracy of statistics such as Krippendorff
$\alpha$ [6]. This is not novel, but it materially governs how
LALM-audio-judge validation should be reported, motivating our
per-regime treatment (Section~\ref{sec:agree}) over a single headline
number.

\section{Study Design}
We rated 209 sessions in total: 152 full-duplex agent-client
conversations spanning 13 accent-and-condition strata (six clean strata
across the accents American-1, American-2, Indian, British, French, and
Italian, plus seven codec-degraded strata that pair accents with
distinct conversational-stress personas), together with 57 adversarial
perturbed clips derived from held-out high-quality baselines. Sessions
were drawn from a production customer-support voice agent. Three
Salesforce-contracted annotators, with no prior exposure to the
product, scored every session on the eight production dimensions
using a 1-5 Likert scale, with rubric anchors visible at scale points
1, 3, and 5 and a structured calibration meeting before the main batch
began. With 8 dimensions scored by 3 raters, this yields
$209 \times 8 \times 3 = 5{,}016$ rating observations.

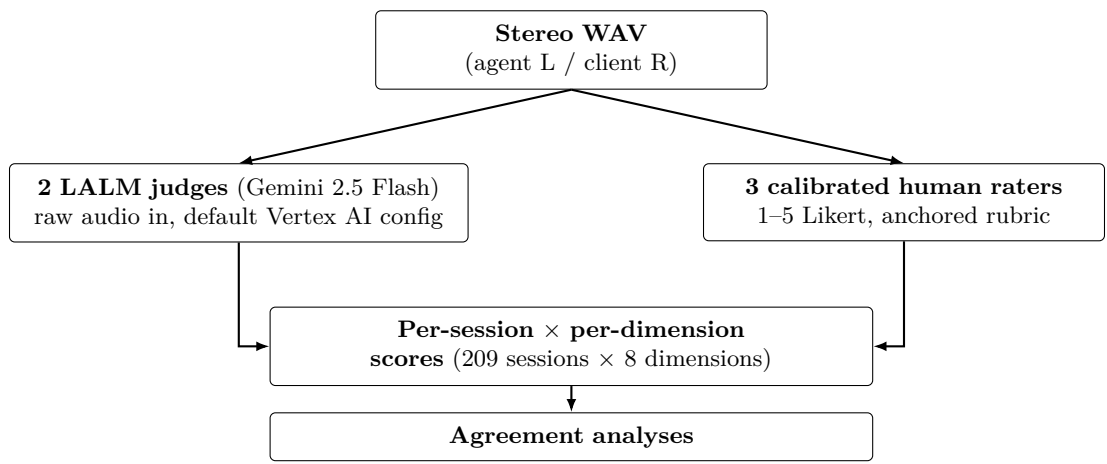
\begin{figure}[!ht]
\centering
\footnotesize
\begin{tikzpicture}[
    box/.style={draw, rounded corners=2pt, align=center, inner sep=5pt,
                font=\footnotesize},
    arr/.style={-{Latex[length=5pt]}, thick},
]
\node[box, text width=1.9in] (audio) at (0,0)
{\textbf{Stereo WAV} (agent L / client R)};

\node[box, text width=2.25in, anchor=north] (lalm) at (-4.4,-1.5)
{\textbf{2 LALM judges} (Gemini 2.5 Flash)\\
raw audio in, default Vertex AI config};

\node[box, text width=1.95in, anchor=north] (human) at (4.4,-1.5)
{\textbf{3 calibrated human raters}\\
1--5 Likert, anchored rubric};

\node[box, text width=3.0in, anchor=north] (scores) at (0,-3.4)
{\textbf{Per-session $\times$ per-dimension scores}
(209 sessions $\times$ 8 dimensions)};

\node[box, text width=3.0in, anchor=north] (analysis) at (0,-4.8)
{\textbf{Agreement analyses}};

\draw[arr] (audio.south) -- (lalm.north);
\draw[arr] (audio.south) -- (human.north);
\draw[arr] (lalm.south) |- (scores.west);
\draw[arr] (human.south) |- (scores.east);
\draw[arr] (scores.south) -- (analysis.north);
\end{tikzpicture}
\caption{Study design. Each stereo session is scored independently by
two production LALM judges (Gemini 2.5 Flash, raw audio input) and by
three calibrated human raters, on the same eight dimensions.}
\label{fig:system}
\end{figure}

We score every session with Gemini 2.5 Flash via the Vertex AI
generate-content API, using the same prompts and JSON schemas already
in production. Both production judges, AgentSpeechFidelity (dimensions
1--4) and ConversationalAudioQuality (dimensions 5--8), receive the raw
stereo WAV audio bytes directly as an audio input part (agent on the
left channel, human client on the right), with no transcription or
intermediate feature extraction; each returns structured JSON scores
under the default Vertex AI generation configuration (temperature 1,
thinking enabled). We compare the LALM ratings against the three human
raters on equal footing, treating the LALM as a candidate fourth rater.
Adversarial-arm clips are scored by the same judges through an
identical scoring pipeline. Figure~\ref{fig:system} shows the overall
design.

Across the study we report four agreement statistics, each defined at
its point of use and specified in full in Appendix~\ref{app:nw} and the
surrounding subsections: pairwise Spearman $\rho$ (with 1{,}000-iteration
bootstrap 95 percent confidence intervals, seed 42) for rank agreement;
the proportion of sessions within 1 point of the three-rater human mean
for simple agreement; ordinal Krippendorff $\alpha$ for chance-corrected
inter-rater reliability; and Newcombe-Wilson hybrid confidence intervals
on the LALM-minus-human recall difference for the adversarial arm.

\begin{figure}[!ht]
\centering
\footnotesize
\begin{tikzpicture}[
    node distance=6pt,
    box/.style={draw, rounded corners=2pt, align=center, inner sep=5pt,
                text width=2.9in, font=\footnotesize},
    sub/.style={draw, rounded corners=2pt, align=center, inner sep=5pt,
                text width=2.55in, font=\footnotesize},
    arr/.style={-{Latex[length=5pt]}, thick},
]
\node[box] (all) {Production customer-support voice agent\\
\textbf{209 rated stereo sessions} $\times$ 8 dimensions $\times$ 3 raters\\
(5{,}016 rating observations)};

\node[sub, below=14pt of all, xshift=-1.55in] (conv)
{\textbf{152 conversational sessions}\\
13 accent-and-condition strata\\
(6 clean + 7 codec-degraded)};

\node[sub, below=14pt of all, xshift=1.55in] (adv)
{\textbf{57 adversarial clips}\\
10 baselines $\times$ 6 DSP defects\\
(60 nominal; 3 failed to render)};

\node[sub, below=12pt of adv] (recall)
{\textbf{Recall analysis:} 8 baselines with\\
full 3-rater coverage; $n \leq 8$ per\\
(defect, dimension) cell, 4 cells $n=7$};

\draw[arr] (all.south) -- (conv.north);
\draw[arr] (all.south) -- (adv.north);
\draw[arr] (adv.south) -- (recall.north);
\end{tikzpicture}
\caption{Sample-flow for the 209-session analysis pool. All 209 sessions
contribute to the per-dimension agreement statistics
(Table~\ref{tab:agreement}); the adversarial arm additionally supports
the defect-detection analysis, which is restricted to the 8 baselines
with complete three-rater coverage.}
\label{fig:flow}
\end{figure}
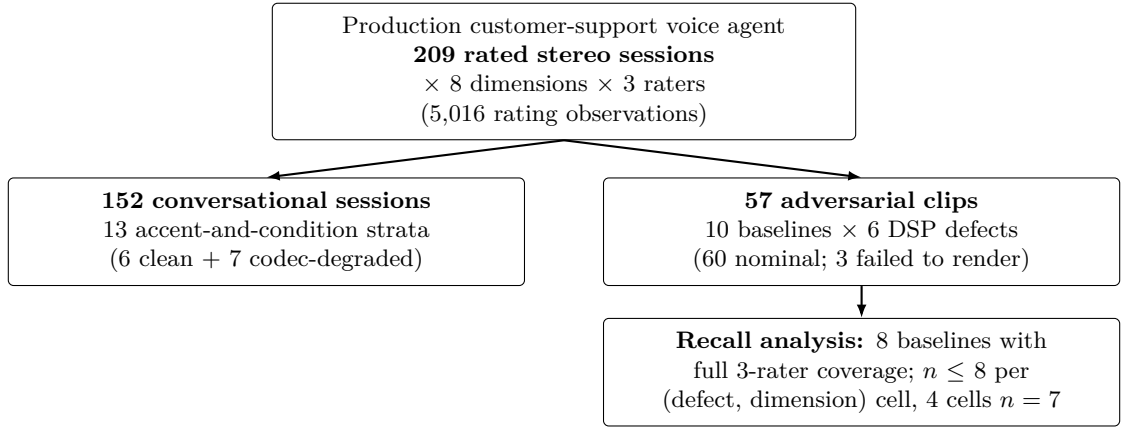

Figure~\ref{fig:flow} summarises how the corpus partitions into the
analysis pools.

\section{Evidence}

\subsection{The LALM agrees with humans about as much as humans agree
with each other}
\label{sec:fourth-rater}
\begin{figure}[!ht]
\centering
\includegraphics[width=5.5in]{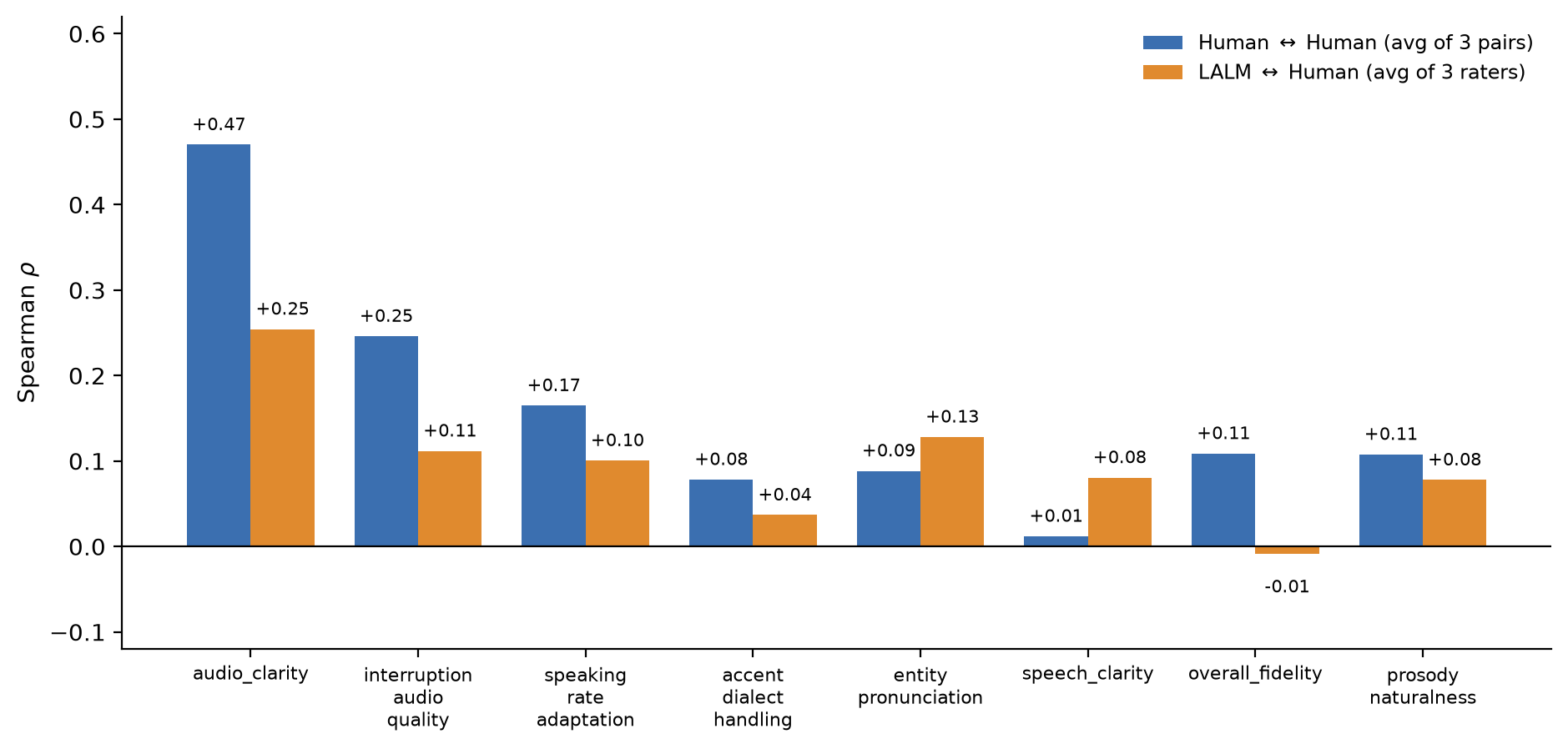}
\caption{Per-dimension Spearman $\rho$. Blue: pairwise human-human
$\rho$ averaged across the three rater pairs. Orange: LALM-human $\rho$
averaged across the three (LALM, human) pairs. Where the two bars are
close, the LALM tracks human ratings on par with a fourth human rater
on that dimension.}
\label{fig:fourth-rater}
\end{figure}

We compute pairwise Spearman $\rho$ between the LALM and each individual
human rater, average the three (LALM, human) values, and compare against
the average of the three pairwise rater-rater Spearman $\rho$ values
among the humans (Figure~\ref{fig:fourth-rater}). On 5 of 8 dimensions
(accent\_dialect\_handling, entity\_pronunciation, speech\_clarity,
prosody\_naturalness, and speaking\_rate\_adaptation), the LALM's
agreement with humans is on par with the agreement humans reach among
themselves: the LALM-human $\rho$ departs from the human-human $\rho$ by
at most 0.07. These are the five tightest-matching dimensions in the
corpus (gaps of 0.03 to 0.07), separated by a clear margin from the
sixth (0.12). On two of them (entity\_pronunciation, speech\_clarity),
the LALM correlates with humans more strongly than the humans correlate
with each other (0.13 vs 0.09 and 0.08 vs 0.01). This test concerns
\emph{rank agreement} only; a dimension can show near-human rank
agreement here yet still have low absolute agreement with the human
mean (Section~\ref{sec:agree}). speaking\_rate\_adaptation is exactly
such a case, and we return to it there.

This parity is reinforced by a distributional test. Under
1{,}000-iteration bootstrap resampling (seed 42), the 95 percent
confidence interval of the LALM-human $\rho$ overlaps that of the
human-human $\rho$ on 7 of 8 dimensions: on those seven, the LALM's
agreement with humans is not statistically separable from the agreement
humans reach with one another. The exception is audio\_clarity, where
the intervals are disjoint (human-human $[+0.34, +0.58]$ versus
LALM-human $[+0.00, +0.30]$), consistent with the gap already visible
in the point estimates. We read this test as supporting rather than a
headline result, since the overlap is driven in part by the wide
confidence intervals that the modest inter-human agreement produces; we
therefore anchor the comparison on the per-dimension gaps above. The
human-human interval is not tabulated elsewhere in this paper; it is
computed with the same bootstrap procedure as Appendix~\ref{app:bootstrap},
resampling sessions rather than rater pairs.

On two further dimensions (audio\_clarity, interruption\_audio\_quality),
the LALM correlates with humans positively but more weakly than humans
correlate with each other. On audio\_clarity specifically, human-human
$\rho$ is 0.47 and LALM-human $\rho$ is 0.25; the LALM correlation is
positive but approximately half the human-human value. On the eighth
dimension (overall\_fidelity) the LALM-human $\rho$ is essentially zero
($-0.01$), while humans correlate among themselves at only 0.11; neither
exhibits a strong rank-discrimination signal.

The implication for deployment is direct: if the deployment criterion is
``does the LALM behave like a fourth human rater?'', the data support
yes on 5 of 8 dimensions, partial yes on 2, and inconclusive on 1.

\subsection{Simple-agreement on production dimensions is high}
\label{sec:agree}
For dimensions where humans saturate at the top of the rating scale,
rank correlation can be uninformative because the underlying score
distribution has very low variance. The complementary statistic is
simple agreement: the proportion of sessions where the LALM rating is
within 1 point of the three-rater human mean. Table~\ref{tab:agreement}
reports this number alongside rank correlation and the mean human
rating.

\begin{table}[!ht]
\centering
\small
\begin{tabular}{@{}lrrrrr@{}}
\toprule
\textbf{Dimension} & \textbf{\% within 1} & \textbf{LALM-human $\rho$} & \textbf{Avg human $\rho$} & \textbf{Krip.\ $\alpha$} & \textbf{Mean human} \\
\midrule
accent\_dialect\_handling      & 92.3\% & $+0.04$ & $+0.08$ & $+0.04$ & 4.88 \\
audio\_clarity                 & 89.0\% & $+0.25$ & $+0.47$ & $+0.03$ & 4.33 \\
speech\_clarity                & 90.9\% & $+0.08$ & $+0.01$ & $-0.15$ & 4.43 \\
entity\_pronunciation          & 66.5\% & $+0.13$ & $+0.09$ & $+0.01$ & 4.71 \\
prosody\_naturalness           & 67.5\% & $+0.08$ & $+0.11$ & $-0.05$ & 3.94 \\
interruption\_audio\_quality   & 60.3\% & $+0.11$ & $+0.25$ & $+0.05$ & 3.20 \\
speaking\_rate\_adaptation     & 48.8\% & $+0.10$ & $+0.17$ & $+0.08$ & 4.03 \\
overall\_fidelity              & 43.5\% & $-0.01$ & $+0.11$ & $-0.01$ & 4.38 \\
\bottomrule
\end{tabular}
\caption{LALM-vs-human-mean agreement statistics on the full $n=209$
3-rater-rated pool (distinct from the adversarial recall analysis in
Section~\ref{sec:defect}, which is restricted to 8 baselines). ``\%
within 1'' is the proportion of sessions where the LALM rating is
within 1 point of the across-rater mean human rating; ``Avg human
$\rho$'' is the average pairwise human-human Spearman $\rho$; ``Krip.\
$\alpha$'' is the chance-corrected inter-rater Krippendorff $\alpha$
with ordinal weighting [6].}
\label{tab:agreement}
\end{table}

Six of eight dimensions show LALM agreement with the human mean of 60
percent or higher, and three reach 89 percent or above. Two dimensions,
speaking\_rate\_adaptation and overall\_fidelity, fall below 50 percent.
We treat these as the dimensions on which LALM ratings are
insufficiently reliable to serve as a standalone metric
(Section~\ref{sec:care}). Note that speaking\_rate\_adaptation appeared
among the five rank-agreement dimensions in Section~\ref{sec:fourth-rater}:
its LALM-human \emph{rank} correlation tracks the human-human level, but
its absolute LALM-vs-mean within-1 agreement is low, so it is
rank-reliable yet not deployable as a standalone metric. The two
criteria measure different things, and speaking\_rate\_adaptation is the
one dimension where they diverge.

The chance-corrected Krippendorff $\alpha$ column exposes why simple
agreement is the right lens for these dimensions. Ordinal $\alpha$ is
near zero or negative on every dimension, ranging only from $-0.15$ to
$+0.08$, even where within-1 agreement exceeds 90 percent, for example
$\alpha = -0.15$ on speech\_clarity despite 90.9 percent within-1
agreement. This is the reliability paradox in its starkest form: the
human panel agrees on the vast majority of sessions in absolute terms,
yet a chance-corrected statistic registers essentially no agreement,
because the rating variance on a saturated dimension is too small for
$\alpha$ to resolve. We therefore report $\alpha$ alongside, not instead
of, simple agreement, and caution that a near-zero $\alpha$ on a
ceiling dimension reflects the shape of the rating distribution rather
than genuine rater disagreement.

\subsection{Adversarial defect detection: asymmetric sensitivity, few
significant cells}
\label{sec:defect}
\begin{figure}[!ht]
\centering
\includegraphics[width=5.0in, height=5.48in]{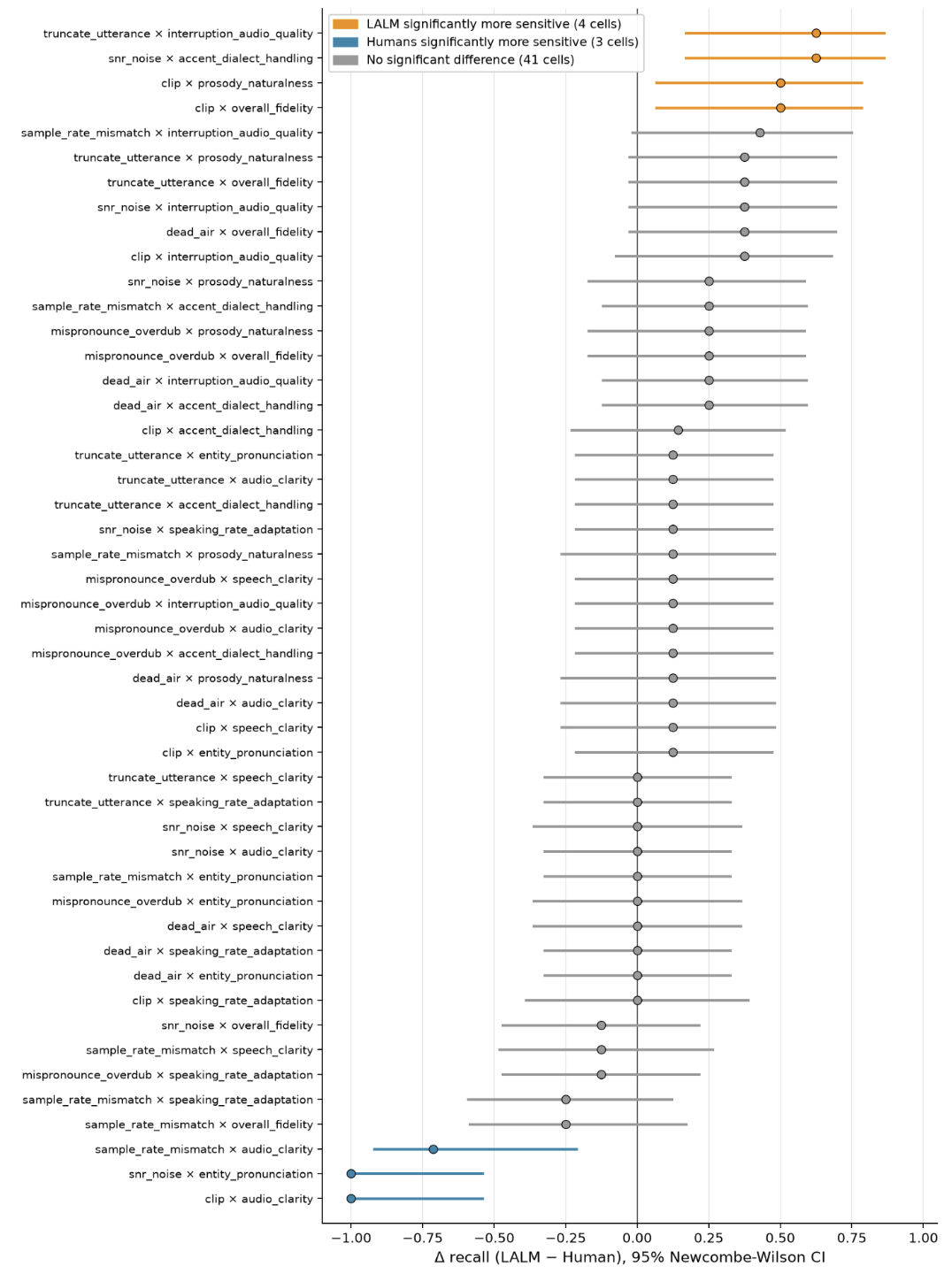}
\caption{Forest plot of LALM-minus-human recall difference across all
48 (defect, dimension) cells. Orange: 4 cells where the LALM is
significantly more sensitive than humans (CI excludes 0 above). Blue:
3 cells where humans are significantly more sensitive. Gray: 41 cells
with no significant difference.}
\label{fig:forest}
\end{figure}

To test sensitivity to controlled audio degradation, we apply six
DSP-level defects (hard amplitude clipping, dead air, additive Gaussian
noise at $-2$ to $-8$~dB SNR, mid-utterance truncation, phoneme-region
overdubbing, and 8~kHz sample-rate down/upsample) to 10
high-baseline-quality sessions, a nominal 60-clip corpus of which 57
rendered successfully. We restrict the recall analysis to the 8
baselines with full three-rater coverage, giving up to $n=8$ per
(defect, dimension) cell (four cells drop to $n=7$). For each cell we
measure the recall: the proportion of perturbed clips on which the
judge's score drops by at least 1 point relative to that clip's
baseline. For the LALM the score is its single rating; for humans it is
the across-rater mean, so ``human recall'' is the fraction of clips on
which the mean human rating dropped by at least 1 point (not a
majority- or any-rater vote). Newcombe-Wilson hybrid 95 percent
confidence intervals on the LALM-minus-human recall difference are
reported in Figure~\ref{fig:forest}.

On 4 of 48 cells the LALM is significantly more sensitive than humans,
on 3 humans are significantly more sensitive, and the remaining 41 show
no significant difference (an absence of power at $n \leq 8$, not
demonstrated equivalence). These flags come from Newcombe-Wilson
intervals that treat the two recalls as independent; because both
judges score the same clips, a paired McNemar re-test is stricter, and
under it only the two most extreme cells remain significant
(Appendix~\ref{app:paired}). Of the 3 cells where the LALM misses
artefacts humans catch, two
fall on audio\_clarity (clipping and sample-rate mismatch), and the
third is signal-to-noise degradation on entity\_pronunciation. The two
audio\_clarity misses are the operationally salient pair, since despite
coming from two different DSP mechanisms (amplitude clipping and
sample-rate aliasing), they concentrate on the same dimension. The
implication for deployment is that for the broad defect-detection use
case, swapping in the LALM does not meaningfully reduce coverage; for
the narrow audio\_clarity blind spot, an additional check is warranted
(Section~\ref{sec:care}).

The most instructive pattern is that the LALM's response to a single
artefact is dimension-dependent. Hard amplitude clipping is a clear
example: under clipping the LALM lowers prosody\_naturalness and
overall\_fidelity on 4 of 8 clips (where humans lower neither, 0 of 8),
yet lowers audio\_clarity on 0 of 8 (where humans lower it on all 8).
The identical waveform corruption is thus registered by the LALM as a
prosody and fidelity problem but not as a clarity problem, while human
raters do the reverse. We read this as evidence that the two judges
route the same acoustic evidence to different rubric constructs: the
LALM appears to treat clipping-induced spectral distortion as unnatural
\emph{delivery} (prosody, overall fidelity), whereas humans categorise
the same distortion as degraded \emph{signal quality} (audio\_clarity).
The pattern recurs for other artefacts, sample-rate mismatch is caught
by humans on audio\_clarity (0.71 recall) but not the LALM (0.00), while
signal-to-noise degradation is caught by the LALM on accent handling,
where humans miss it entirely (0.00 recall); on interruption quality
the same defect is caught by both judges, though the LALM catches more
(1.00 vs.\ 0.625). The operational
consequence is that LALM and human coverage are complementary rather
than nested: a session flagged by neither is more trustworthy than one
flagged by either alone, which argues for retaining a human spot-check
tier on the specific (artefact, dimension) pairs where the LALM is
blind, rather than for a uniform confidence adjustment.

\subsection{The LALM is architecturally more consistent than human
raters}
\label{sec:determinism}
A deployment-relevant property of an LLM judge is intra-judge
consistency. Both judges ran under the default Vertex AI generation
configuration (temperature 1, thinking enabled) for this study, so we
make no determinism claim from configuration alone; even a
temperature-0 setting would not guarantee exactly reproducible output
run to run [11]. The more robust argument is architectural: an
automated judge has no analogue to a human rater's fatigue, mood, or
drifting rubric interpretation over a multi-week batch, so it is
structurally more consistent than a human panel independent of the
exact sampling configuration. Human raters offer no such guarantee.
Between-rater
variation in our pool is substantial: rater means on the same dimension
differ by more than a full point on some dimensions, by 1.11 points on
speech\_clarity (3.74 vs 4.85) and 1.12 points on audio\_clarity (3.66
vs 4.78). We stress that this is a stability property, not an accuracy
property: this observation speaks to reproducibility of the headline
metric rather than to correctness, which the preceding agreement tests
address. A related observation is
that LALM-vs-human-mean agreement (Table~\ref{tab:agreement}) is in many
cases as high as the agreement between any individual human-rater pair,
because the LALM tends toward the central tendency the human panel
collectively expresses. For production monitoring, where a stable and
reproducible headline number is valuable in its own right, this is a
useful operating point.

\subsection{Decomposition: agreement on natural conversations alone}
\label{sec:natural}
The agreement pool in Table~\ref{tab:agreement} combines 152 natural
full-duplex conversations with the 57 adversarial perturbed clips.
Table~\ref{tab:natural} recomputes the same statistics on the 152
natural conversations alone, the population a deployed judge actually
scores, and \textbf{the headline results hold}: the LALM-human $\rho$
still lands within 0.07 of the human-human $\rho$ on 5 of 8 dimensions,
and within-1 agreement still exceeds 60 percent on 6 of 8. Neither
top-line claim depends on the adversarial clips. One dimension, however,
shifts substantially, and the shift is itself informative. On the pooled data, audio\_clarity
looked like an \emph{informative} dimension where humans rank-agree
strongly (human-human $\rho = 0.47$) and the LALM trails (0.25); on
natural conversations alone, the human-human $\rho$ falls to 0.07 and
the LALM-human $\rho$ to 0.04. The pooled correlation was therefore
measuring shared sensitivity to the injected defects, not rank
agreement on natural audio, on which humans themselves barely
rank-discriminate. Restricted to natural conversations, no dimension
exhibits strong human rank-discrimination (the highest human-human
$\rho$ is 0.27, on interruption\_audio\_quality); the LALM is a
defensible fourth rater precisely in the near-ceiling regime that
dominates real traffic, and there is correspondingly no
strong-rank-discrimination dimension for it to fall short on.

\begin{table}[!ht]
\centering
\small
\begin{tabular}{@{}lrrrrr@{}}
\toprule
\textbf{Dimension} & \textbf{\% within 1} & \textbf{LALM-human $\rho$} & \textbf{Avg human $\rho$} & \textbf{Krip.\ $\alpha$} & \textbf{Mean human} \\
\midrule
accent\_dialect\_handling      & 94.1\% & $+0.07$ & $+0.08$ & $+0.05$ & 4.86 \\
audio\_clarity                 & 92.8\% & $+0.04$ & $+0.07$ & $-0.29$ & 4.59 \\
speech\_clarity                & 92.8\% & $+0.02$ & $+0.01$ & $-0.13$ & 4.47 \\
entity\_pronunciation          & 63.8\% & $+0.12$ & $+0.12$ & $+0.07$ & 4.73 \\
prosody\_naturalness           & 69.7\% & $+0.06$ & $+0.14$ & $+0.02$ & 3.92 \\
interruption\_audio\_quality   & 60.5\% & $+0.12$ & $+0.27$ & $+0.05$ & 3.05 \\
speaking\_rate\_adaptation     & 59.9\% & $+0.10$ & $+0.11$ & $+0.03$ & 3.88 \\
overall\_fidelity              & 40.8\% & $-0.06$ & $+0.20$ & $+0.12$ & 4.39 \\
\bottomrule
\end{tabular}
\caption{Agreement statistics on the 152 natural conversations only
(the adversarial arm excluded), directly comparable to
Table~\ref{tab:agreement}. The two headline counts (5 of 8 within 0.07;
6 of 8 above 60 percent within-1) are unchanged; audio\_clarity's
human-human $\rho$ falls from 0.47 to 0.07 once the engineered
variation of the adversarial clips is removed.}
\label{tab:natural}
\end{table}

\subsection{Cross-model replication}
\label{sec:cross-model}
Every result so far uses a single model, Gemini 2.5 Flash. To check
whether the agreement pattern is a property of the approach or an
artefact of that one model, we re-scored all 209 sessions with the
same two judges, changing nothing but the underlying LALM, against two
further Gemini models from a newer model family: Gemini 3.5 Flash and
Gemini 3.1 Pro Preview. Both re-scoring runs completed with zero
errors across all 418 (session, judge) calls.

Gemini 3.5 Flash reproduces the paper's central finding, with a mixed
comparison to Gemini 2.5 Flash: within-1 agreement with the human mean
improves to 8 of 8 dimensions (versus 6 of 8), while the rank-correlation
gap stays within 0.07 of the human-human baseline on 4 of 8 dimensions,
one fewer than Gemini 2.5 Flash's 5 of 8, with the remaining four at
0.08--0.15 except audio\_clarity, which trails on every model tested
(Appendix \ref{app:cross-model}). The result we would most want a second model
to confirm, that the LALM is a defensible fourth rater on most
dimensions, holds again on an independently trained, later-generation
model, which is the strongest evidence in this paper that the finding
is about the approach rather than about one checkpoint.

Gemini 3.1 Pro Preview tells a more qualified story: its rank
correlations are comparable to the other two models, but its absolute
scores run substantially below the human mean on several dimensions,
most severely audio\_clarity ($-1.52$ points) and
speaking\_rate\_adaptation ($-1.44$ points), which pulls its within-1
agreement below 60 percent on three dimensions. This is a calibration
offset rather than a ranking failure, and we return to its deployment
implication in Section~\ref{sec:calibration}.

\begin{table}[!ht]
\centering
\small
\begin{tabular}{@{}lrr@{}}
\toprule
\textbf{Model} & \textbf{Rank-corr.\ within 0.07} & \textbf{Within-1 $\geq$60\%} \\
\midrule
gemini-2.5-flash       & 5 of 8 & 6 of 8 \\
gemini-3.5-flash       & 4 of 8 & 8 of 8 \\
gemini-3.1-pro-preview & 5 of 8 & 5 of 8 \\
\bottomrule
\end{tabular}
\caption{Headline agreement counts across all three models tested.
Rank-correlation transfers consistently (4--5 of 8 dimensions within
the 0.07 gap threshold for every model); simple agreement is where
gemini-3.1-pro-preview diverges, driven by the calibration offset
detailed above rather than a ranking failure. Full per-dimension
results in Appendix~\ref{app:cross-model}.}
\label{tab:cross-model-summary}
\end{table}

\section{Where Deployment Requires Care}
\label{sec:care}
Four areas of the result space ask for additional handling before
deployment is recommended without qualification.

\subsection{audio\_clarity rank-discrimination gap}
The rank-discrimination gap on audio\_clarity identified in
Section~\ref{sec:fourth-rater} lives almost entirely in the
\emph{degraded}-audio regime supplied by the adversarial clips
(Section~\ref{sec:natural}): on natural conversations both correlations
are near zero. That makes the gap operationally relevant exactly where it
matters, when audio quality is actually compromised. The LALM is still
positively correlated and agrees on 89 percent of sessions within 1
point, so the deployment risk is not that LALM ratings are wrong on
average, but that they are less reliable on the relative ranking of two
sessions of comparable but imperfect audio. Routing audio\_clarity into
a human-pair sample for any session where the LALM rating is below 4
would resolve this without losing the cost benefit on the 89 percent of
sessions where the LALM agrees with humans cleanly.

\subsection{Clipping artefacts on audio\_clarity specifically}
On the cell clip $\times$ audio\_clarity, humans unanimously detect the
artefact (8 of 8) and the LALM unanimously fails to (0 of 8). This is a
directional miss large enough to matter if audio\_clarity is judged
exclusively by the LALM and amplitude clipping is a known production
artefact. A lightweight DSP detector for hard amplitude clipping (which
is cheap to compute on the raw audio buffer) could route any session
where clipping is detected into a human review queue, leaving the LALM
as the headline judge on the remaining sessions. This is one of only
two cells in the entire 48-cell study robust to the paired McNemar
re-test (Appendix~\ref{app:paired}); the other is signal-to-noise
degradation on entity\_pronunciation (Section~\ref{sec:defect}), which
warrants an analogous safeguard for that dimension. A second
audio\_clarity miss, sample-rate mismatch, is directional (0 of 7 LALM
vs.\ 5 of 7 human) but does not survive the paired test.

\subsection{Two contentious dimensions need rubric work, not LALM work}
On overall\_fidelity and speaking\_rate\_adaptation, neither humans nor
the LALM produce reliable rank-correlation signals: human-human $\rho$
is 0.11 and 0.17, LALM-human $\rho$ is $-0.01$ and 0.10. These
dimensions are not saturated at the top of the scale the way
accent\_dialect\_handling is (median rating 4.33 and 4.00 respectively,
versus a median of 5.00 there), despite mean ratings of 4.38 and 4.03;
humans simply disagree about what the rubrics mean. The
blocker on these dimensions is rubric construct validity, not the LALM.
We recommend dropping them from headline deployment metrics while a
rubric reformulation pass is underway, and then re-running this
validation post-revision to test whether anchor sharpening drives them
into the cleanly-deployable regime.

\subsection{Re-check calibration, not just ranking, after a model swap}
\label{sec:calibration}
Section~\ref{sec:cross-model} found that rank-ordering ability
transferred cleanly across two further Gemini models, but
absolute calibration did not: Gemini 3.1 Pro Preview rates
audio\_clarity and speaking\_rate\_adaptation more than a full Likert
point below the human mean on average, which alone drops those two
dimensions' within-1 agreement to 34.9 percent and 37.3 percent
respectively (Appendix Table~\ref{tab:cross-model-31}), well below the
60 percent bar this paper otherwise treats as a usability floor. A
team that swaps the underlying model expecting the validation to carry
over should not assume it does: rank-correlation checks alone would
have missed this, since Gemini 3.1 Pro Preview's rank correlation gap
is within 0.07 of the human-human baseline on 5 of 8 dimensions,
including speaking\_rate\_adaptation, one of the two dimensions where
its calibration is furthest off, and matches Gemini 2.5 Flash while
exceeding Gemini 3.5 Flash on this measure
(Table~\ref{tab:cross-model-summary}). We
recommend re-running the simple-agreement check in
Section~\ref{sec:agree} specifically, on a small session sample,
before promoting any model swap to production, rather than relying on
rank-correlation parity as a proxy for calibration parity.

\section{Cost and Operational Implications}
This is a directional internal estimate, not a reproducible unit-cost
model. Under conservative human-side
assumptions, a calibrated rater needs 6 to 8 minutes to score a
5-minute session across all eight dimensions, so three-rater coverage
of a 100-session weekly batch is roughly 35 person-hours, about 0.88 of
a fully-loaded engineering FTE. The LALM alternative is two API calls
per session (one per judge), a few dollars per week dominated by audio
input tokens. We do not claim a precise ratio, but the gap between roughly
one FTE of rater time and a few-dollar API bill is of order two
magnitudes, even after reserving full human review for the two
contentious dimensions and a 10 percent spot-check on the rest.

The throughput unlock matters more than the dollar saving. Three-rater
human coverage caps how many sessions we can audit per week; with the
LALM as primary judge and humans as a spot-check and escalation tier,
that cap shifts from rater hours to engineering attention, making a
1{,}000-session weekly cadence feasible without added rater capacity.

\section{Limitations of This Evidence}
Five caveats deserve naming explicitly. First, the primary evidence
base is a single LALM (Gemini 2.5 Flash) on a single production
customer-support agent; generalisation to other agents would still
warrant a fresh validation. We partially address model generalisation
directly (Section~\ref{sec:cross-model}): rank-ordering ability
replicated on two further Gemini models, but calibration did not
on one of them (Section~\ref{sec:calibration}), so a model swap should
be treated as requiring its own simple-agreement check, not assumed to
inherit this validation's headline numbers. Second, our human reference
standard rests on only three
raters; a larger panel would tighten the human-human agreement
estimates and better generalise our conclusions; a larger-rater-pool
remains the cleanest validity test. Third, the adversarial
arm's $n=7$--8 per (defect, dimension) cell is too small for
Newcombe-Wilson CIs to resolve close cells; the 41 non-significant
cells reflect a lack of power, not demonstrated equivalence, so the
``matches'' half of the 45-of-48 figure means ``no detectable
difference at this sample size,'' not proven parity. Fourth, the
48-cell adversarial analysis has no multiple-comparisons correction;
roughly 2 to 3 of the 7 flagged cells could be false positives at 95
percent, so we treat them as hypotheses to confirm, not established
effects, relying on the aggregate pattern rather than any single cell.
Fifth, LALM and human recall are paired, not independent
as Newcombe-Wilson assumes; an exact paired McNemar re-test
(Appendix~\ref{app:paired}) confirms only the two most extreme cells of
seven. None of these caveats change the
recommendation that the evidence supports phased deployment with the
safeguards in Section~\ref{sec:care}.

\section{Conclusion}
On a 209-session corpus (152 full-duplex agent conversations across 13
strata and 57 adversarial defect-injected clips), Gemini 2.5 Flash
matches the rank-correlation level of the human-rater pool on five of
eight dimensions, agrees with the across-rater human mean within 1
point on 60 to 92 percent of sessions on six of eight dimensions, and
in the adversarial arm is significantly more defect-sensitive than
humans on 4 of 48 (defect, dimension) cells and less sensitive on 3,
with no detectable difference on the remaining 41 at this sample size. These are the standard agreement statistics
that LALM-as-judge validation studies are scored on, and on each of
them the evidence supports deploying the LALM as a substitute or fourth
rater, provided the safeguards in Section~\ref{sec:care} are in place.
A cross-model check (Section~\ref{sec:cross-model}) found that this
pattern is not an artefact of the specific model: rank-ordering ability
held on two further Gemini models, though one of them required a
calibration correction that rank-correlation alone would not have
surfaced (Section~\ref{sec:calibration}), which we take as a reason to
verify calibration explicitly after any future model swap rather than
as a reason to doubt the underlying approach. The approximate two-orders-of-magnitude
operational cost reduction and the throughput unlock that follow are
downstream consequences of this validity claim, not arguments for it.

\clearpage
\section*{References}
\addcontentsline{toc}{section}{References}

\input{appendix}

\end{document}

%% file: appendix.tex

\appendix
\clearpage

\section*{Appendix}
\addcontentsline{toc}{section}{Appendix}
\phantomsection

\renewcommand{\thesection}{\Alph{section}}
\setcounter{section}{0}

\section{Study Design Details}

\subsection{Full 14-stratum breakdown}
\label{app:strata}

Table~\ref{tab:strata} lists every stratum in the analysis pool, with
its sample size, accent, persona, codec setting, and background-noise
setting. The 14 rows comprise 13 conversational strata (152 sessions)
and the adversarial arm (57 perturbed clips), summing to the full
209-session pool. The Italian clean condition contributes a partial
five-session batch (\texttt{baseline-italian-5}); it is retained in the
pool but is smaller than the other clean strata because collection was
cut short.

\begin{table}[!ht]
\centering
\small
\begin{tabular}{@{}lllccr@{}}
\toprule
\textbf{Config} & \textbf{Accent} & \textbf{Persona} & \textbf{Codec} & \textbf{Noise} & \textbf{n} \\
\midrule
clean-american-1        & American-1 & neutral            & off & off & 10 \\
clean-american-2        & American-2 & neutral            & off & off & 38 \\
clean-indian            & Indian     & neutral            & off & off & 10 \\
clean-british           & British    & neutral            & off & off & 10 \\
clean-french            & French     & neutral            & off & off & 10 \\
baseline-italian-5      & Italian    & neutral            & off & off &  5 \\
\midrule
codec-american-1        & American-1 & barge-in           & on  & off &  9 \\
codec-american-2        & American-2 & dual-conversation  & on  & off & 10 \\
codec-indian            & Indian     & slow-speech        & on  & off & 10 \\
codec-indian-2          & Indian     & rapid-speech       & on  & off & 10 \\
codec-british           & British    & whisper            & on  & off & 10 \\
codec-french            & French     & fragmented-speech  & on  & off & 10 \\
codec-italian           & Italian    & steamroll          & on  & off & 10 \\
\midrule
adversarial             & American-1 & neutral (perturbed) & baseline & \textit{DSP} & 57 \\
\bottomrule
\end{tabular}
\caption{Full stratum breakdown of the 209-session analysis pool (152
conversational sessions across 13 strata plus 57 adversarial perturbed
clips). \textit{Persona} identifies the simulated-client behavioural
mode used for that cell.}
\label{tab:strata}
\end{table}

The adversarial arm applies six perturbations to each of 10
high-baseline-quality sessions, a nominal 60-clip corpus; 57 clips
rendered successfully (3 perturbations failed to generate), which is
the count shown for the adversarial row in Table~\ref{tab:strata}. The
recall analysis is restricted to the 8 baselines with full three-rater
coverage on both the baseline and its perturbed siblings, yielding up
to $n=8$ per (defect, dimension) cell. Four cells drop to $n=7$: recall
requires a LALM baseline score, a LALM perturbed score, and the human
rating for the same (defect, dimension), and in these four cells one of
those components is missing for a single clip (the clip itself rendered
and was rated on the other dimensions), so that clip is excluded from
the affected cell only.

\subsection{Rater calibration protocol}
\label{app:calibration}

Three Salesforce-contracted annotators, with no prior exposure to the
product, served on the panel and are referred to as \texttt{R1},
\texttt{R2}, \texttt{R3} throughout the paper and this appendix.
Calibration followed a two-stage protocol:

\begin{itemize}[leftmargin=1.4em, itemsep=1pt]
\item \textit{Round 1 (individual)}: each rater independently scored
the same 5-session calibration set on all 8 dimensions using the
rubric anchors listed in Appendix~\ref{app:rubrics}, without seeing
each other's ratings.
\item \textit{Meeting}: the three raters and a moderator reviewed
disagreements dimension by dimension, discussed cases where the
rubric was ambiguous, and agreed on interpretation. No rubric
anchors were changed post-meeting; interpretation notes were captured
informally.
\item \textit{Round 2 (main batch)}: raters scored the full
209-session analysis pool, which comprises the 152 conversational
sessions and the 57 adversarial perturbed clips, using the aligned
rubric interpretation. Rating order was randomised across raters to
avoid correlated fatigue effects.
\end{itemize}

Raters scored sessions through an internal eVerse annotation tool
(built for this evaluation workflow, distinct from the Labelforce
labeling product). Raters were blind to which sessions were natural
conversations versus adversarial perturbed clips, and were never shown
the LALM's scores at any point before or during their own rating.

\subsection{Rubric anchors}
\label{app:rubrics}

The rubrics below are abbreviated for space, showing the anchor
descriptions at scale points 1, 3, and 5; the full rubric text (all
five points, with entity-type lists and worked examples) is embedded in
the LALM prompt reproduced in Appendix~\ref{app:prompts} and released in
the supplementary archive. Raters may use 2 and 4 as intermediate
values. Rating on a 5-point Likert scale with anchor
descriptions follows the general protocol of ITU-T Recommendation
P.808 [5], adapted here for full-duplex two-channel conversation
audio.

\paragraph{entity\_pronunciation}
\begin{itemize}[leftmargin=1.4em, itemsep=0pt]
\item \textbf{5 (Perfect)}: All entities pronounced clearly, correctly,
and distinctly.
\item \textbf{3 (Acceptable)}: 2--3 minor issues or one moderate issue
(e.g.\ wrong digit, unclear letter).
\item \textbf{1 (Very Poor)}: Critical entities are incomprehensible or
majorly incorrect.
\end{itemize}

\paragraph{speech\_clarity}
\begin{itemize}[leftmargin=1.4em, itemsep=0pt]
\item \textbf{5}: Every word is perfectly clear and easy to understand.
\item \textbf{3}: Some words unclear but overall understandable.
\item \textbf{1}: Speech is often incomprehensible or heavily distorted.
\end{itemize}

\paragraph{prosody\_naturalness}
\begin{itemize}[leftmargin=1.4em, itemsep=0pt]
\item \textbf{5 (Indistinguishable from human)}: Natural pitch, rhythm,
stress, and emotional expressiveness.
\item \textbf{3 (Acceptable)}: Recognisably synthetic but acceptable;
somewhat robotic.
\item \textbf{1 (Very robotic)}: Monotone, mechanical, uncanny-valley.
\end{itemize}

\paragraph{overall\_fidelity}
\textit{The production LALM judge emits this dimension on a 0--2 scale,
which we rescale to 1--5 (0$\to$1, 1$\to$3, 2$\to$5) for comparison with
the other seven dimensions; the LALM therefore occupies only the three
values $\{1,3,5\}$ on this dimension. Human raters, by contrast, use the
full 1--5 scale. The overall\_fidelity agreement figures thus compare a
coarse three-point LALM grid against five-point human ratings, and
should be read with that granularity mismatch in mind; it is one reason
we classify overall\_fidelity as a contentious dimension on which LALM
ratings are not deployable as a standalone metric.}
\begin{itemize}[leftmargin=1.4em, itemsep=0pt]
\item \textbf{2 (High)}: All critical entities correct, clear speech,
natural prosody, no major quality issues.
\item \textbf{1 (Medium)}: Most entities correct; clear speech with
minor artefacts; prosody acceptable.
\item \textbf{0 (Low)}: Critical entity errors, clarity issues that
impair comprehension, or very robotic prosody.
\end{itemize}

\paragraph{interruption\_audio\_quality}
\begin{itemize}[leftmargin=1.4em, itemsep=0pt]
\item \textbf{5}: Seamless overlaps, no pops/clicks/clipping.
\item \textbf{3}: Minor audio artefacts at transitions but understandable.
\item \textbf{1}: Severe audio corruption during interruptions.
\end{itemize}

\paragraph{speaking\_rate\_adaptation}
\begin{itemize}[leftmargin=1.4em, itemsep=0pt]
\item \textbf{5}: Agent perfectly matches caller's pace.
\item \textbf{3}: Fixed agent pace that is acceptable but does not adapt.
\item \textbf{1}: Major pace mismatch that makes conversation awkward.
\end{itemize}

\paragraph{accent\_dialect\_handling}
\begin{itemize}[leftmargin=1.4em, itemsep=0pt]
\item \textbf{5}: Conversation flows naturally regardless of accent;
appropriate style match.
\item \textbf{3}: Acceptable conversation despite accent differences.
\item \textbf{1}: Accent mismatch causes confusion.
\end{itemize}

\paragraph{audio\_clarity}
\begin{itemize}[leftmargin=1.4em, itemsep=0pt]
\item \textbf{5}: Crystal clear audio, natural silences, consistent
volume.
\item \textbf{3}: Acceptable quality with minor background noise or
volume issues.
\item \textbf{1}: Poor audio quality, excessive noise, inconsistent
volume, dead air.
\end{itemize}

\subsection{LALM prompts and JSON schemas}
\label{app:prompts}

Two prompts are used, matching the two production judges:
\texttt{AgentSpeechFidelity} (dimensions 1--4) and
\texttt{ConversationalAudioQuality} (dimensions 5--8). Both are shown
to Gemini 2.5 Flash with the stereo WAV attached as an audio part.
This section reproduces the JSON output schemas verbatim and
paraphrases the prompt structure; the full, unedited prompt text for
both judges is released in the supplementary archive as
\texttt{speech\_fidelity\_prompt.py} and
\texttt{conversational\_audio\_quality.py}.

\paragraph{Common preamble (paraphrased from both prompts)}
Both judges are told the WAV is a stereo full-duplex recording with the
AI agent on the LEFT channel and the human caller on the RIGHT channel.
The judges are asked to focus on the AGENT channel for
\texttt{AgentSpeechFidelity} and on the two-party interaction for
\texttt{ConversationalAudioQuality}. Both return structured JSON
matching the schemas below.

\paragraph{AgentSpeechFidelity output schema (verbatim)}
\begin{verbatim}
{
  "entity_pronunciation": {
    "score": <int 1-5>,
    "entities_evaluated": [
      {"text": <str>, "type": <str>,
       "correct": <bool>, "notes": <str>}
    ],
    "issues": [<str>],
    "justification": <str>
  },
  "speech_clarity": {
    "score": <int 1-5>,
    "justification": <str>
  },
  "prosody_naturalness": {
    "score": <int 1-5>,
    "justification": <str>
  },
  "overall_fidelity": {
    "rating": <int 0-2>,
    "reasoning": <str>
  }
}
\end{verbatim}

\paragraph{ConversationalAudioQuality output schema (verbatim, with
the intended range noted)}
Unlike \texttt{AgentSpeechFidelity}, the \texttt{ConversationalAudioQuality}
JSON schema constrains \texttt{score} to type integer only; it does not
enforce a minimum or maximum in the schema itself. The 1--5 range is
required by the prompt and rubric, not by schema validation.
\begin{verbatim}
{
  "interruption_audio_quality":
    {"score": <int>, "justification": <str>},
  "speaking_rate_adaptation":
    {"score": <int>, "justification": <str>},
  "accent_dialect_handling":
    {"score": <int>, "justification": <str>},
  "audio_clarity":
    {"score": <int>, "justification": <str>}
}
\end{verbatim}
Prompt/rubric-intended range: 1--5 for all four dimensions.

Both prompts embed the same rubric anchors listed in
Appendix~\ref{app:rubrics}. The full prompt source files are included
in the reproducibility manifest (Appendix~\ref{app:manifest}).

\section{Full Numerical Results}

\subsection{Complete per-dimension agreement statistics}
\label{app:per-dim}

Table~\ref{tab:full-per-dim} reports every agreement statistic used in
the paper on the full 209-session pool. Bootstrap 95\% CIs on
LALM-human $\rho$ use 1{,}000 iterations with session resampling.
Source: \texttt{h1\_per\_dim.csv} and \texttt{lalm\_as\_4th\_rater.csv}.

\begin{table}[!ht]
\centering
\footnotesize
\setlength{\tabcolsep}{4pt}
\begin{tabular}{@{}lrrrrrr@{}}
\toprule
\textbf{Dimension} & \textbf{$\rho_{HH}$} & \textbf{$\rho_{LH}$} & \textbf{CI} & \textbf{\%$\leq$1} & \textbf{L\%1} & \textbf{Mean H} \\
\midrule
audio\_clarity                & $+0.47$ & $+0.25$ & $[+0.00, +0.30]$ & 67.0\% & 89.0\% & 4.33 \\
interruption\_audio\_quality  & $+0.25$ & $+0.11$ & $[+0.03, +0.29]$ & 17.7\% & 60.3\% & 3.20 \\
speaking\_rate\_adaptation    & $+0.17$ & $+0.10$ & $[+0.02, +0.26]$ & 54.5\% & 48.8\% & 4.03 \\
accent\_dialect\_handling     & $+0.08$ & $+0.04$ & $[-0.12, +0.17]$ & 88.5\% & 92.3\% & 4.88 \\
entity\_pronunciation         & $+0.09$ & $+0.13$ & $[+0.04, +0.32]$ & 76.6\% & 66.5\% & 4.71 \\
speech\_clarity               & $+0.01$ & $+0.08$ & $[+0.04, +0.33]$ & 50.7\% & 90.9\% & 4.43 \\
overall\_fidelity             & $+0.11$ & $-0.01$ & $[-0.13, +0.15]$ & 62.2\% & 43.5\% & 4.38 \\
prosody\_naturalness          & $+0.11$ & $+0.08$ & $[+0.03, +0.28]$ & 52.2\% & 67.5\% & 3.94 \\
\bottomrule
\end{tabular}
\caption{Per-dimension agreement on $n=209$ sessions.
$\rho_{HH}$ is the average pairwise human-human Spearman $\rho$;
$\rho_{LH}$ is the average LALM-human $\rho$ across the three
(LALM, rater) pairs; \textbf{CI} is the bootstrap 95\% CI on
$\rho_{LH}$; \textbf{\%$\leq$1} is the three-rater max-min spread
$\leq 1$; \textbf{L\%1} is the LALM-vs-human-mean within-1
proportion.}
\label{tab:full-per-dim}
\end{table}

\subsection{Cross-model replication: gemini-3.5-flash and
gemini-3.1-pro-preview}
\label{app:cross-model}

To check whether the agreement pattern in
Table~\ref{tab:full-per-dim} is specific to Gemini 2.5 Flash or holds
across model choice, we re-scored all 209 sessions with both judges
(AgentSpeechFidelity, ConversationalAudioQuality) unchanged, swapping
only the underlying model. Two further Gemini models from a newer
model family were available to us at the time of writing:
\texttt{gemini-3.5-flash} and \texttt{gemini-3.1-pro-preview}. Both
were called via the Vertex AI \texttt{generateContent} API at each
model's default sampling configuration. All 209 sessions $\times$ 2
judges $=$ 418 calls per model completed with zero errors for both
models.

Table~\ref{tab:cross-model-35} reports Gemini 3.5 Flash's full
per-dimension results, directly comparable to
Table~\ref{tab:full-per-dim}. Simple agreement improves on the
original: 8 of 8 dimensions exceed 60\% within-1 agreement (versus 6 of
8 for Gemini 2.5 Flash). The rank-correlation gap is within 0.07 on 4
of 8 dimensions, one fewer than Gemini 2.5 Flash's 5 of 8, though the
gaps on the dimensions that miss the threshold are still modest
(0.08--0.15 on all but audio\_clarity). audio\_clarity remains the one
dimension where the LALM-human $\rho$ trails the human-human $\rho$ by
a wide margin (0.26), consistent with Section~\ref{sec:natural}'s
finding that this gap is a property of the adversarial-arm audio
regime, not of any one model.

\begin{table}[!ht]
\centering
\footnotesize
\setlength{\tabcolsep}{4pt}
\begin{tabular}{@{}lrrrrrrr@{}}
\toprule
\textbf{Dimension} & \textbf{$\rho_{HH}$} & \textbf{$\rho_{LH}$} & \textbf{Gap} & \textbf{L\%1} & \textbf{Mean L} & \textbf{Mean H} & \textbf{Bias} \\
\midrule
accent\_dialect\_handling      & $+0.08$ & $+0.12$ & 0.04 & 95.7\% & 4.53 & 4.88 & $-0.35$ \\
audio\_clarity                 & $+0.47$ & $+0.21$ & 0.26 & 82.3\% & 4.14 & 4.33 & $-0.20$ \\
speech\_clarity                & $+0.01$ & $+0.09$ & 0.08 & 94.3\% & 4.74 & 4.43 & $+0.30$ \\
entity\_pronunciation          & $+0.09$ & $+0.18$ & 0.09 & 81.8\% & 4.57 & 4.71 & $-0.14$ \\
prosody\_naturalness           & $+0.11$ & $+0.13$ & 0.02 & 92.3\% & 3.86 & 3.94 & $-0.08$ \\
interruption\_audio\_quality   & $+0.25$ & $+0.40$ & 0.15 & 70.3\% & 3.74 & 3.20 & $+0.55$ \\
speaking\_rate\_adaptation     & $+0.17$ & $+0.20$ & 0.04 & 69.4\% & 3.28 & 4.03 & $-0.75$ \\
overall\_fidelity              & $+0.11$ & $+0.14$ & 0.03 & 74.2\% & 4.36 & 4.38 & $-0.02$ \\
\bottomrule
\end{tabular}
\caption{Per-dimension agreement for gemini-3.5-flash on the same
$n=209$ session pool, directly comparable to
Table~\ref{tab:full-per-dim}. $\rho_{HH}$ is unchanged from
Table~\ref{tab:full-per-dim} (same human raters, same sessions);
\textbf{Gap} is $|\rho_{LH} - \rho_{HH}|$; \textbf{L\%1} is the
LALM-vs-human-mean within-1 proportion; \textbf{Bias} is the mean
LALM score minus the mean human score (negative: LALM rates lower
than humans on average). Source: \texttt{h1\_per\_dim\_gemini35flash.csv}.}
\label{tab:cross-model-35}
\end{table}

Table~\ref{tab:cross-model-31} reports the same statistics for
Gemini 3.1 Pro Preview. Rank correlation is comparable to the other
two models (5 of 8 dimensions within the 0.07 gap threshold), but
simple agreement is weaker on several dimensions because the model's
mean score runs substantially below the human mean: by $-1.52$ points
on audio\_clarity, $-1.44$ on speaking\_rate\_adaptation, and
$-0.80$ or lower on three further dimensions
(overall\_fidelity, accent\_dialect\_handling, entity\_pronunciation).
Three dimensions
(audio\_clarity, speaking\_rate\_adaptation, overall\_fidelity) fall
below the 60\% within-1 threshold as a direct consequence, most
severely audio\_clarity at 34.9\%, where humans and Gemini 2.5 Flash
both rate the same sessions 4--5 on average (Table~\ref{tab:full-per-dim})
while Gemini 3.1 Pro Preview rates them 2.81 on average. This is a
calibration offset, not a ranking failure: the model still orders
sessions in roughly the same relative order as humans on this
dimension ($\rho_{LH} = +0.20$, comparable to Gemini 2.5 Flash's
$+0.25$ and Gemini 3.5 Flash's $+0.21$); it simply anchors its scale
lower. Section~\ref{sec:care} discusses the deployment implication.

\begin{table}[!ht]
\centering
\footnotesize
\setlength{\tabcolsep}{4pt}
\begin{tabular}{@{}lrrrrrrr@{}}
\toprule
\textbf{Dimension} & \textbf{$\rho_{HH}$} & \textbf{$\rho_{LH}$} & \textbf{Gap} & \textbf{L\%1} & \textbf{Mean L} & \textbf{Mean H} & \textbf{Bias} \\
\midrule
accent\_dialect\_handling      & $+0.08$ & $+0.17$ & 0.09 & 76.6\% & 4.08 & 4.88 & $-0.80$ \\
audio\_clarity                 & $+0.47$ & $+0.20$ & 0.27 & 34.9\% & 2.81 & 4.33 & $-1.52$ \\
speech\_clarity                & $+0.01$ & $+0.03$ & 0.01 & 86.1\% & 4.68 & 4.43 & $+0.25$ \\
entity\_pronunciation          & $+0.09$ & $+0.08$ & 0.00 & 61.2\% & 3.91 & 4.71 & $-0.80$ \\
prosody\_naturalness           & $+0.11$ & $+0.16$ & 0.05 & 67.0\% & 3.39 & 3.94 & $-0.55$ \\
interruption\_audio\_quality   & $+0.25$ & $+0.41$ & 0.17 & 71.8\% & 3.34 & 3.20 & $+0.15$ \\
speaking\_rate\_adaptation     & $+0.17$ & $+0.21$ & 0.05 & 37.3\% & 2.59 & 4.03 & $-1.44$ \\
overall\_fidelity              & $+0.11$ & $+0.12$ & 0.01 & 52.6\% & 3.54 & 4.38 & $-0.85$ \\
\bottomrule
\end{tabular}
\caption{Per-dimension agreement for gemini-3.1-pro-preview on the
same $n=209$ session pool, directly comparable to
Table~\ref{tab:full-per-dim} and Table~\ref{tab:cross-model-35}. Column
definitions match Table~\ref{tab:cross-model-35}. Source:
\texttt{h1\_per\_dim\_gemini31propreview.csv}.}
\label{tab:cross-model-31}
\end{table}

Table~\ref{tab:cross-model-summary} in the main text (Section~\ref{sec:cross-model})
summarises the headline counts across all three models.

\subsection{Complete 48-cell defect-recall table}
\label{app:defects}

Table~\ref{tab:defects} is the full data behind the forest plot in
Figure~\ref{fig:forest} of the main paper. \textit{L rec} and \textit{H rec} are the
LALM and human recall (proportion of perturbed clips where the
respective judge's score drops by $\geq 1$ point relative to
baseline); \textit{$\Delta$} is L rec $-$ H rec; \textit{Sig} flags
cells whose Newcombe-Wilson 95\% CI on $\Delta$ excludes zero.
Source: \texttt{h3\_defect\_recall\_with\_ci.csv}.

\begin{table}[!ht]
\centering
\footnotesize
\setlength{\tabcolsep}{3.5pt}
\begin{tabular}{@{}llrrrrl@{}}
\toprule
\textbf{Defect} & \textbf{Dimension} & \textbf{n} & \textbf{L rec} & \textbf{H rec} & \textbf{$\Delta$} & \textbf{Sig} \\
\midrule
clip & accent\_dialect\_handling & 7 & 0.14 & 0.00 & $+0.14$ &  \\
clip & audio\_clarity & 8 & 0.00 & 1.00 & $-1.00$ & H$>$L \\
clip & entity\_pronunciation & 8 & 0.12 & 0.00 & $+0.12$ &  \\
clip & interruption\_audio\_quality & 8 & 0.50 & 0.12 & $+0.38$ &  \\
clip & overall\_fidelity & 8 & 0.50 & 0.00 & $+0.50$ & L$>$H \\
clip & prosody\_naturalness & 8 & 0.50 & 0.00 & $+0.50$ & L$>$H \\
clip & speaking\_rate\_adaptation & 7 & 0.14 & 0.14 & $\phantom{+}0.00$ &  \\
clip & speech\_clarity & 8 & 0.25 & 0.12 & $+0.12$ &  \\
\midrule
dead\_air & accent\_dialect\_handling & 8 & 0.25 & 0.00 & $+0.25$ &  \\
dead\_air & audio\_clarity & 8 & 0.25 & 0.12 & $+0.12$ &  \\
dead\_air & entity\_pronunciation & 8 & 0.00 & 0.00 & $\phantom{+}0.00$ &  \\
dead\_air & interruption\_audio\_quality & 8 & 0.25 & 0.00 & $+0.25$ &  \\
dead\_air & overall\_fidelity & 8 & 0.38 & 0.00 & $+0.38$ &  \\
dead\_air & prosody\_naturalness & 8 & 0.25 & 0.12 & $+0.12$ &  \\
dead\_air & speaking\_rate\_adaptation & 8 & 0.00 & 0.00 & $\phantom{+}0.00$ &  \\
dead\_air & speech\_clarity & 8 & 0.12 & 0.12 & $\phantom{+}0.00$ &  \\
\midrule
mispronounce\_overdub & accent\_dialect\_handling & 8 & 0.12 & 0.00 & $+0.12$ &  \\
mispronounce\_overdub & audio\_clarity & 8 & 0.12 & 0.00 & $+0.12$ &  \\
mispronounce\_overdub & entity\_pronunciation & 8 & 0.12 & 0.12 & $\phantom{+}0.00$ &  \\
mispronounce\_overdub & interruption\_audio\_quality & 8 & 0.12 & 0.00 & $+0.12$ &  \\
mispronounce\_overdub & overall\_fidelity & 8 & 0.38 & 0.12 & $+0.25$ &  \\
mispronounce\_overdub & prosody\_naturalness & 8 & 0.38 & 0.12 & $+0.25$ &  \\
mispronounce\_overdub & speaking\_rate\_adaptation & 8 & 0.00 & 0.12 & $-0.12$ &  \\
mispronounce\_overdub & speech\_clarity & 8 & 0.12 & 0.00 & $+0.12$ &  \\
\midrule
sample\_rate\_mismatch & accent\_dialect\_handling & 8 & 0.25 & 0.00 & $+0.25$ &  \\
sample\_rate\_mismatch & audio\_clarity & 7 & 0.00 & 0.71 & $-0.71$ & H$>$L \\
sample\_rate\_mismatch & entity\_pronunciation & 8 & 0.00 & 0.00 & $\phantom{+}0.00$ &  \\
sample\_rate\_mismatch & interruption\_audio\_quality & 7 & 0.43 & 0.00 & $+0.43$ &  \\
sample\_rate\_mismatch & overall\_fidelity & 8 & 0.12 & 0.38 & $-0.25$ &  \\
sample\_rate\_mismatch & prosody\_naturalness & 8 & 0.25 & 0.12 & $+0.12$ &  \\
sample\_rate\_mismatch & speaking\_rate\_adaptation & 8 & 0.00 & 0.25 & $-0.25$ &  \\
sample\_rate\_mismatch & speech\_clarity & 8 & 0.12 & 0.25 & $-0.12$ &  \\
\midrule
snr\_noise & accent\_dialect\_handling & 8 & 0.62 & 0.00 & $+0.62$ & L$>$H \\
snr\_noise & audio\_clarity & 8 & 1.00 & 1.00 & $\phantom{+}0.00$ &  \\
snr\_noise & entity\_pronunciation & 8 & 0.00 & 1.00 & $-1.00$ & H$>$L \\
snr\_noise & interruption\_audio\_quality & 8 & 1.00 & 0.62 & $+0.38$ &  \\
snr\_noise & overall\_fidelity & 8 & 0.88 & 1.00 & $-0.12$ &  \\
snr\_noise & prosody\_naturalness & 8 & 0.88 & 0.62 & $+0.25$ &  \\
snr\_noise & speaking\_rate\_adaptation & 8 & 0.12 & 0.00 & $+0.12$ &  \\
snr\_noise & speech\_clarity & 8 & 0.88 & 0.88 & $\phantom{+}0.00$ &  \\
\midrule
truncate\_utterance & accent\_dialect\_handling & 8 & 0.12 & 0.00 & $+0.12$ &  \\
truncate\_utterance & audio\_clarity & 8 & 0.12 & 0.00 & $+0.12$ &  \\
truncate\_utterance & entity\_pronunciation & 8 & 0.12 & 0.00 & $+0.12$ &  \\
truncate\_utterance & interruption\_audio\_quality & 8 & 0.62 & 0.00 & $+0.62$ & L$>$H \\
truncate\_utterance & overall\_fidelity & 8 & 0.38 & 0.00 & $+0.38$ &  \\
truncate\_utterance & prosody\_naturalness & 8 & 0.38 & 0.00 & $+0.38$ &  \\
truncate\_utterance & speaking\_rate\_adaptation & 8 & 0.00 & 0.00 & $\phantom{+}0.00$ &  \\
truncate\_utterance & speech\_clarity & 8 & 0.00 & 0.00 & $\phantom{+}0.00$ &  \\
\bottomrule
\end{tabular}
\caption{Defect $\times$ dimension recall table (48 cells). Positive
$\Delta$ means the LALM is more sensitive to the defect than the
human panel; negative means the humans are.}
\label{tab:defects}
\end{table}

Of 48 cells, 4 are significantly $\Delta > 0$ (L$>$H), 3 are
significantly $\Delta < 0$ (H$>$L), and 41 have CIs that include
zero.

\subsection{Per-rater breakdown}
\label{app:per-rater}

Table~\ref{tab:per-rater} shows each rater's mean per dimension and
each pairwise rater-rater Spearman $\rho$. Rater labels are
anonymised as \texttt{R1}, \texttt{R2}, \texttt{R3}. Source:
\texttt{h1\_per\_dim.csv}.

\begin{table}[!ht]
\centering
\footnotesize
\setlength{\tabcolsep}{4pt}
\begin{tabular}{@{}lrrrrrr@{}}
\toprule
\textbf{Dimension} & \textbf{R1 mean} & \textbf{R2 mean} & \textbf{R3 mean} & \textbf{$\rho_{12}$} & \textbf{$\rho_{13}$} & \textbf{$\rho_{23}$} \\
\midrule
audio\_clarity                & 3.66 & 4.78 & 4.56 & $+0.45$ & $+0.42$ & $+0.54$ \\
interruption\_audio\_quality  & 3.43 & 2.24 & 3.92 & $+0.20$ & $+0.13$ & $+0.41$ \\
speaking\_rate\_adaptation    & 4.00 & 4.44 & 3.66 & $+0.12$ & $+0.12$ & $+0.25$ \\
accent\_dialect\_handling     & 4.95 & 4.98 & 4.70 & $-0.02$ & $+0.17$ & $+0.09$ \\
entity\_pronunciation         & 4.37 & 4.85 & 4.91 & $+0.11$ & $+0.14$ & $+0.01$ \\
speech\_clarity               & 3.74 & 4.71 & 4.85 & $-0.03$ & $+0.12$ & $-0.05$ \\
overall\_fidelity             & 3.83 & 4.68 & 4.64 & $+0.11$ & $+0.08$ & $+0.13$ \\
prosody\_naturalness          & 3.56 & 4.60 & 3.67 & $+0.07$ & $+0.19$ & $+0.05$ \\
\bottomrule
\end{tabular}
\caption{Per-rater means and pairwise rater-rater Spearman $\rho$
across the 209-session pool. \texttt{R1}, \texttt{R2}, \texttt{R3}
are anonymous rater identifiers.}
\label{tab:per-rater}
\end{table}

Rater-mean spread on the same dimension reaches 1.11 points on
speech\_clarity (R1 3.74 vs R3 4.85) and 1.12 points on
audio\_clarity (R1 3.66 vs R2 4.78). This variance underlies the
observation in Section~\ref{sec:determinism} that the LALM's
consistent ratings land closer to the across-rater mean than any
single rater's ratings do to another rater's on many dimensions.

\subsection{Per-stratum breakdown}
\label{app:per-stratum}

Table~\ref{tab:per-stratum} summarises LALM-vs-human agreement per
stratum for the two most illustrative dimensions:
\texttt{audio\_clarity}, where humans show real rank-discrimination
(Section~\ref{sec:fourth-rater}), and \texttt{accent\_dialect\_handling},
which sits near the top of the scale (Section~\ref{sec:natural}). The
full 14-stratum $\times$ 8-dimension table is available in
\texttt{h2\_per\_config\_per\_dim.csv}.

\begin{table}[!ht]
\centering
\footnotesize
\setlength{\tabcolsep}{4pt}
\begin{tabular}{@{}lrrrrr@{}}
\toprule
& & \multicolumn{2}{c}{\textbf{audio\_clarity}} & \multicolumn{2}{c}{\textbf{accent\_dialect\_handling}} \\
\cmidrule(lr){3-4}\cmidrule(lr){5-6}
\textbf{Stratum} & \textbf{n} & \textbf{$\rho_{L\bar{H}}$} & \textbf{L\%1} & \textbf{$\rho_{L\bar{H}}$} & \textbf{L\%1} \\
\midrule
clean-american-1     & 10 & $+0.09$ & 100 & --- & 100 \\
clean-american-2     & 38 & $+0.21$ & 95 & $-0.14$ & 95 \\
clean-indian         & 10 & $-0.54$ & 90 & $+0.60$ & 90 \\
clean-british        & 10 & $-0.06$ & 90 & $-0.11$ & 100 \\
clean-french         & 10 & $-0.32$ & 90 & $+0.68$ & 90 \\
codec-american-1     &  9 & --- & 100 & --- & 100 \\
codec-american-2     & 10 & $-0.07$ & 100 & --- & 100 \\
codec-indian         & 10 & $-0.11$ & 90 & $-0.41$ & 90 \\
codec-indian-2       & 10 & $-0.27$ & 80 & $-0.10$ & 80 \\
codec-british        & 10 & $+0.07$ & 90 & --- & 100 \\
codec-french         & 10 & $+0.51$ & 100 & $+0.78$ & 100 \\
codec-italian        & 10 & $+0.04$ & 100 & $-0.27$ & 100 \\
adversarial          & 57 & $+0.56$ & 79 & $-0.18$ & 88 \\
\bottomrule
\end{tabular}
\caption{Per-stratum LALM-vs-human agreement on two representative
dimensions. Here $\rho_{L\bar{H}}$ is the Spearman correlation between
the LALM and the \emph{mean} human rating (distinct from $\rho_{LH}$ in
Table~\ref{tab:full-per-dim}, which averages the three LALM-vs-individual
-rater correlations); per-stratum $n$ is too small to compute the
three-pair average reliably. Small per-cell $n$ (typically 10) inflates
the CI on per-stratum $\rho$; per-stratum numbers should be treated as
indicative rather than definitive. A dash (---) marks cells where
$\rho$ is undefined because the LALM assigned a constant rating within
the stratum (zero rank variance).}
\label{tab:per-stratum}
\end{table}

\subsection{Score-region partition}
\label{app:regions}

Table~\ref{tab:regions} partitions each dimension by mean-human
rating into ceiling ($\bar{H} \geq 4$), midrange
($2.5 \leq \bar{H} < 4$), and below ($\bar{H} < 2.5$) buckets and
re-computes LALM-vs-mean-human $\rho$ within each region. On most
dimensions, the below-ceiling bucket is empty because raters
rarely use the bottom of the scale; the two dimensions with
substantive below-ceiling samples show sharply different
behaviour, discussed in the main text (Sections~\ref{sec:fourth-rater}
and~\ref{sec:determinism}). Source: \texttt{e\_ceiling\_collapse.csv}.

\begin{table}[!ht]
\centering
\footnotesize
\setlength{\tabcolsep}{4pt}
\begin{tabular}{@{}lrrrrrr@{}}
\toprule
& \multicolumn{2}{c}{\textbf{Ceiling ($\bar{H}\geq 4$)}} & \multicolumn{2}{c}{\textbf{Midrange}} & \multicolumn{2}{c}{\textbf{Below ($\bar{H}<2.5$)}} \\
\cmidrule(lr){2-3}\cmidrule(lr){4-5}\cmidrule(lr){6-7}
\textbf{Dimension} & \textbf{n} & \textbf{$\rho_{L\bar{H}}$} & \textbf{n} & \textbf{$\rho_{L\bar{H}}$} & \textbf{n} & \textbf{$\rho_{L\bar{H}}$} \\
\midrule
entity\_pronunciation        & 191 & $+0.23$ &  18 & $+0.36$ &   0 & --- \\
speech\_clarity              & 185 & $+0.04$ &  24 & $+0.07$ &   0 & --- \\
prosody\_naturalness         & 125 & $+0.03$ &  82 & $-0.05$ &   2 & --- \\
overall\_fidelity            & 176 & $+0.01$ &  32 & $-0.53$ &   1 & --- \\
interruption\_audio\_quality &  56 & $+0.08$ & 105 & $+0.01$ &  48 & $-0.06$ \\
speaking\_rate\_adaptation   & 139 & $+0.21$ &  69 & $+0.19$ &   1 & --- \\
accent\_dialect\_handling    & 203 & $+0.00$ &   6 & --- &     0 & --- \\
audio\_clarity               & 182 & $-0.03$ &  17 & $-0.37$ &  10 & $+0.67$ \\
\bottomrule
\end{tabular}
\caption{LALM-human Spearman $\rho$ partitioned by mean-human rating
region. Empty cells indicate the region has zero eligible sessions
for that dimension.}
\label{tab:regions}
\end{table}

\subsection{Persona vs codec effect deltas}
\label{app:persona-codec}

Table~\ref{tab:persona} decomposes the codec-degradation-arm rating
shifts into their persona and codec components. \textit{Codec$_{\mathrm{avg}}$}
is the mean rating change from codec on/off averaging over both accents;
\textit{Persona} is the codec-on rating difference between
\texttt{codec-american-1} (barge-in) and \texttt{codec-american-2}
(dual-conversation), which share the broad American accent family and
codec setting but differ in persona; we note the persona effect is not
fully isolated from the American-1/American-2 accent difference, so
this decomposition is indicative rather than a clean persona contrast.
\textit{Neutral baseline} is the same-persona same-codec across-accent
delta as a chance floor. Source: \texttt{d\_personality\_vs\_codec.csv}.

\begin{table}[!ht]
\centering
\footnotesize
\setlength{\tabcolsep}{4pt}
\begin{tabular}{@{}lrrr@{}}
\toprule
\textbf{Dimension} & \textbf{Codec$_{\mathrm{avg}}$} & \textbf{Persona} & \textbf{Neutral} \\
\midrule
entity\_pronunciation        & $-0.16$ & $+0.73$ & $-0.05$ \\
speech\_clarity              & $-0.17$ & $+0.08$ & $+0.05$ \\
prosody\_naturalness         & $-0.13$ & $-0.47$ & $-0.12$ \\
overall\_fidelity            & $+0.09$ & $-0.02$ & $-0.20$ \\
interruption\_audio\_quality & $+0.15$ & $-0.70$ & $+0.25$ \\
speaking\_rate\_adaptation   & $-0.24$ & $-0.38$ & $-0.09$ \\
accent\_dialect\_handling    & $\phantom{+}0.00$ & $-0.04$ & $+0.04$ \\
audio\_clarity               & $-0.15$ & $-0.12$ & $-0.05$ \\
\midrule
\textbf{$|\cdot|$ mean}      & \textbf{0.14} & \textbf{0.32} & \textbf{0.11} \\
\bottomrule
\end{tabular}
\caption{Persona-vs-codec effect decomposition. Persona
effect $/$ codec effect ratio averaged across the eight dimensions
is 2.3; both exceed the neutral baseline of 0.11.}
\label{tab:persona}
\end{table}

\section{DSP Defect Implementations}

\subsection{Parameter table}
\label{app:dsp-params}

Table~\ref{tab:dsp-params} lists the parameter values used for each of
the six adversarial defects. Each defect is applied to an isolated
copy of the baseline WAV; the perturbed clip is then re-registered as
a synthetic run in the evaluation backend and scored by both the LALM
and the three human raters.

\begin{table}[!ht]
\centering
\footnotesize
\begin{tabular}{@{}p{1.5in}p{4in}@{}}
\toprule
\textbf{Defect} & \textbf{Parameter values} \\
\midrule
clip &
Hard amplitude clipping at $\pm 0.85$ of peak amplitude
applied to the agent (LEFT) channel; both channels' peaks are
preserved before clipping. Duration unchanged. \\
\midrule
dead\_air &
A single random 200--800~ms silence inserted into the agent
channel at a uniformly-random position between 20\% and 80\% of
the session duration. Client channel unmodified in the same
time window. \\
\midrule
snr\_noise &
Additive Gaussian white noise scaled to yield SNR of $-2$ to $-8$~dB
(uniformly sampled per session) relative to the loudest 500~ms
window of the agent channel. Applied to both channels. \\
\midrule
sample\_rate\_mismatch &
Downsample the agent channel from 16~kHz to 8~kHz then upsample
back to 16~kHz using linear interpolation; introduces
characteristic aliasing artefacts. Duration and RMS preserved. \\
\midrule
truncate\_utterance &
A single 150--400~ms segment of the agent channel is removed at
a uniformly-random position between 30\% and 70\% of the
session duration. The tail is spliced back with a 5~ms linear
crossfade. Session duration reduced by the truncation length. \\
\midrule
mispronounce\_overdub &
A short vowel-region segment (100--300~ms) in the agent channel
is replaced with a length-matched vowel from a different phoneme
drawn from a small candidate pool. Duration unchanged. \\
\bottomrule
\end{tabular}
\caption{Adversarial defect parameters. All defects operate on the
16 kHz stereo WAV baselines and preserve total duration except where
noted.}
\label{tab:dsp-params}
\end{table}

\subsection{Reference implementation}
\label{app:dsp-code}

The six defects are implemented as pure-Python functions taking a
NumPy array of the stereo signal and returning a modified array of
the same shape (except \texttt{truncate\_utterance}, which returns a
shorter array). Illustrative implementations for the four
signal-domain defects (the two temporal-splicing defects are more
verbose) are:

\begin{verbatim}
import numpy as np

SR = 16000
AGENT = 0  # LEFT channel (agent); RIGHT (1) is the client

def clip(x, thr=0.85):
    """Hard-clip the agent channel to a fraction of its peak."""
    y = x.copy()
    peak = np.max(np.abs(y[:, AGENT]))
    y[:, AGENT] = np.clip(y[:, AGENT], -thr * peak, thr * peak)
    return y

def snr_noise(x, snr_db):
    """Add Gaussian noise at a target SNR (dB), relative to the
    loudest 500 ms window of the agent channel; noise on both."""
    agent = x[:, AGENT]
    win = int(0.5 * SR)
    energy = np.convolve(agent ** 2, np.ones(win), mode="valid")
    start = int(np.argmax(energy))
    peak_rms = np.sqrt(np.mean(agent[start:start + win] ** 2))
    noise_rms = peak_rms / (10 ** (snr_db / 20))
    return x + np.random.normal(0, noise_rms, size=x.shape)

def sample_rate_mismatch(x):
    """8 kHz down/up-sample of the agent channel (linear interp)."""
    y = x.copy()
    agent = y[:, AGENT]
    down = agent[np.arange(0, len(agent), 2)]
    y[:, AGENT] = np.interp(np.arange(len(agent)),
                            np.arange(len(down)) * 2, down)
    return y

def dead_air(x, start_s, dur_s):
    """Zero-fill a window on the agent channel."""
    y = x.copy()
    s, e = int(start_s * SR), int((start_s + dur_s) * SR)
    y[s:e, AGENT] = 0.0
    return y
\end{verbatim}

Truncation and mispronounce-overdub are implemented as
splice-and-crossfade over source-and-donor phoneme regions; the full
implementations are available in the reproducibility manifest
(Appendix~\ref{app:manifest}).

\section{Statistical Methodology}

\subsection{Krippendorff $\alpha$ with ordinal weighting}
\label{app:alpha}

We compute $\alpha$ with the ordinal difference function of
Krippendorff [6], in which the disagreement between two ordinal ratings
is a rank-based squared distance defined over the cumulative marginal
category frequencies (not the raw squared difference $(v_1 - v_2)^2$,
which is the interval metric). We use the reference \texttt{krippendorff}
implementation. Values are reported in the main paper's Table~1 as the
``Krip.\ $\alpha$'' column. We warn readers that on ceiling-effect
dimensions the observed disagreement approaches the expected chance
disagreement, driving $\alpha$ toward zero, or slightly negative,
regardless of the high simple agreement on the same dimension.

\subsection{Newcombe-Wilson hybrid CIs on recall difference}
\label{app:nw}

For each (defect, dimension) cell we form the LALM recall
$\hat{p}_L = k_L / n$ and human recall $\hat{p}_H = k_H / n$, where
$k_L$ and $k_H$ are the counts of perturbed clips where the
respective judge's score drops by $\geq 1$ point relative to
baseline and $n$ is the number of three-rater-covered perturbed
clips for that cell. We compute the 95\% CI on
$\hat{p}_L - \hat{p}_H$ using Newcombe's Method 10 (hybrid Wilson),
which for two independent proportions combines the Wilson score CIs
on each proportion into a lower bound $L$ and upper bound $U$ on the
difference (see Newcombe [4], eqs.\ 12--14). A
cell is flagged \textit{L$>$H} when $L > 0$ and \textit{H$>$L} when
$U < 0$.

\subsection{Paired-test robustness check}
\label{app:paired}

Newcombe's Method 10 treats the LALM and human recalls as two
independent proportions, but they are in fact paired: both judges score
the same perturbed clips. As a robustness check we re-tested each cell
with an exact paired McNemar test on the per-clip discordant pairs (the
clips one judge flags and the other does not). Under this stricter
paired criterion, 2 of the 48 cells reach $p < 0.05$, both in the
Human$>$LALM direction and both perfect 8-of-8-versus-0-of-8 splits:
clip $\times$ audio\_clarity and snr\_noise $\times$ entity\_pronunciation.
The other five cells the Newcombe-Wilson intervals flagged do not
survive the paired test at $n \leq 8$: three of the five have a 5-of-5
(or 5-of-7) discordant split, i.e.\ every discordant pair favors the
same judge, but at this sample size an exact binomial test needs a
perfectly unanimous split to clear $p < 0.05$, so these read as
underpowered directional misses rather than disconfirmed effects. This
reinforces the main-text caution that the individual significant cells
are hypotheses to confirm rather than established effects; only the
two most extreme misses are robust to the paired analysis.

\subsection{Bootstrap parameters}
\label{app:bootstrap}

Confidence intervals on Spearman $\rho$ between the LALM and each
rater use non-parametric bootstrap with the following parameters:
\begin{itemize}[leftmargin=1.4em, itemsep=0pt]
\item Iterations: 1{,}000
\item Resampling unit: session (i.e.\ all 8 dimensions of a
resampled session travel together)
\item Sampling: with replacement
\item CI method: percentile (2.5\%, 97.5\%)
\item Random seed: 42 (documented in the analysis script)
\end{itemize}

\subsection{Human-human $\rho$ confidence intervals}
\label{app:hh-ci}

Section~\ref{sec:fourth-rater} compares the bootstrap 95\% CI of the
LALM-human $\rho$ (Table~\ref{tab:full-per-dim}) against the CI of the
human-human $\rho$, which is not tabulated elsewhere in this paper.
Table~\ref{tab:hh-ci} reports it, computed with the identical
session-level resampling procedure described above, applied to the
average of the three pairwise rater-rater correlations. On 7 of 8
dimensions the two intervals overlap; on audio\_clarity they do not,
which is why that dimension's rank-correlation gap is treated as the
one exception to the LALM-as-fourth-rater parity result. Source:
\texttt{rho\_hh\_ci.csv}.

\begin{table}[!ht]
\centering
\footnotesize
\begin{tabular}{@{}lrrrl@{}}
\toprule
\textbf{Dimension} & \textbf{$\rho_{HH}$} & \textbf{95\% CI ($\rho_{HH}$)} & \textbf{95\% CI ($\rho_{LH}$)} & \textbf{Overlap} \\
\midrule
accent\_dialect\_handling      & $+0.08$ & $[-0.01, +0.29]$ & $[-0.12, +0.17]$ & yes \\
audio\_clarity                 & $+0.47$ & $[+0.34, +0.58]$ & $[+0.00, +0.30]$ & \textbf{no} \\
speech\_clarity                & $+0.01$ & $[-0.05, +0.08]$ & $[+0.04, +0.33]$ & yes \\
entity\_pronunciation          & $+0.09$ & $[-0.00, +0.21]$ & $[+0.04, +0.32]$ & yes \\
prosody\_naturalness           & $+0.11$ & $[+0.02, +0.20]$ & $[+0.03, +0.28]$ & yes \\
interruption\_audio\_quality   & $+0.25$ & $[+0.16, +0.34]$ & $[+0.03, +0.29]$ & yes \\
speaking\_rate\_adaptation     & $+0.17$ & $[+0.08, +0.26]$ & $[+0.02, +0.26]$ & yes \\
overall\_fidelity              & $+0.11$ & $[+0.02, +0.20]$ & $[-0.13, +0.15]$ & yes \\
\bottomrule
\end{tabular}
\caption{Bootstrap 95\% confidence intervals on the human-human
$\rho$ (average of the three pairwise rater correlations) alongside
the LALM-human $\rho$ CI already reported in Table~\ref{tab:full-per-dim}.
Both use the same 1{,}000-iteration, session-level, seed-42 procedure.}
\label{tab:hh-ci}
\end{table}

\subsection{Pre-registration status}
\label{app:prereg}

None of the analyses in this paper were pre-registered. The
three-regime taxonomy, the LALM-as-fourth-rater test, and the
adversarial-defect design were specified before data collection.
The persona-vs-codec decomposition and the per-rater-idiosyncrasy
observation were identified post-hoc during analysis; the paper
treats them as observational. The specific 0.07 rank-correlation-gap
threshold used to report ``5 of 8 dimensions'' in
Section~\ref{sec:fourth-rater} was likewise chosen after inspecting
the data (it falls in an observed gap between the fifth- and
sixth-closest dimensions, 0.07 versus 0.12) rather than fixed in
advance; Table~\ref{tab:full-per-dim} reports the full per-dimension
gaps so a reader can apply a different threshold if preferred.

\section{Reproducibility Manifest}
\label{app:manifest}

\paragraph{Analysis artefacts.}
Every table in this paper is derived from one of the thirteen CSV
files listed below. File names refer to the layout of the
supplementary archive that accompanies this paper on submission.

\begin{center}
\footnotesize
\begin{tabular}{@{}ll@{}}
\toprule
\textbf{File} & \textbf{Sourced tables} \\
\midrule
\texttt{h1\_per\_dim.csv}                & Tables~\ref{tab:full-per-dim}, \ref{tab:per-rater} \\
\texttt{h1\_per\_dim\_natural.csv}       & Table~\ref{tab:natural} \\
\texttt{h1\_per\_dim\_gemini35flash.csv} & Table~\ref{tab:cross-model-35} \\
\texttt{h1\_per\_dim\_gemini31propreview.csv} & Table~\ref{tab:cross-model-31} \\
\texttt{cross\_model\_comparison.csv}    & Table~\ref{tab:cross-model-summary} \\
\texttt{rho\_hh\_ci.csv}                 & Table~\ref{tab:hh-ci} \\
\texttt{lalm\_as\_4th\_rater.csv}        & Table~\ref{tab:full-per-dim}, Fig.~\ref{fig:fourth-rater} \\
\texttt{h2\_per\_config\_per\_dim.csv}   & Table~\ref{tab:per-stratum} \\
\texttt{h3\_defect\_recall\_with\_ci.csv}& Table~\ref{tab:defects}, Fig.~\ref{fig:forest} \\
\texttt{e\_ceiling\_collapse.csv}        & Table~\ref{tab:regions} \\
\texttt{d\_personality\_vs\_codec.csv}   & Table~\ref{tab:persona} \\
\texttt{full\_per\_session.csv}          & raw per-session ratings \\
\texttt{h3\_defect\_recall\_3rater.csv}  & pre-CI defect counts \\
\bottomrule
\end{tabular}
\end{center}

\paragraph{Prompt and schema source files.}
The LALM prompts (Appendix~\ref{app:prompts}) and their JSON schemas
are reproduced verbatim from two production source files:
\texttt{speech\_fidelity\_prompt.py} (AgentSpeechFidelity) and
\texttt{conversational\_audio\_quality.py}
(ConversationalAudioQuality). Both files are included in the
supplementary archive.

\paragraph{Data availability.}
The anonymised CSVs, LALM prompts, JSON schemas, analysis scripts,
and rendered figures described above are publicly available as a
Hugging Face Dataset at
\url{https://huggingface.co/datasets/armaan-sayyad/lalm-judge-validation-full-duplex}
[7], released under CC-BY-4.0 for the data and figures and Apache
License 2.0 for the scripts and prompts. Rater identifiers in the raw CSVs have been anonymised to R1,
R2, R3 and session identifiers have been replaced with stable
hashes. Every headline number in the paper can be reproduced from
the released CSVs using the accompanying scripts. The 60 perturbed
adversarial clips (short, derivative WAVs) will be included in a
subsequent v0.2 release pending final data-clearance review. Raw
session recordings from the primary 209-session corpus will not be
publicly released, as they contain production agent output governed
by internal data-handling policy. Requests for restricted access
from bona fide researchers will be considered on a case-by-case
basis.

\section{Notation Glossary}

\begin{itemize}[leftmargin=1.4em, itemsep=1pt]
\item \textbf{LALM}: Large Audio Language Model (Gemini 2.5 Flash in
this study).
\item \textbf{$\rho_{HH}$}: pairwise human-human Spearman rank
correlation, averaged across the three rater pairs.
\item \textbf{$\rho_{LH}$}: LALM-human Spearman rank correlation,
averaged across the three (LALM, rater) pairs. Used in Appendix
Table~\ref{tab:full-per-dim} and the LALM-as-fourth-rater comparison.
\item \textbf{$\rho_{L\bar{H}}$}: Spearman correlation between the LALM
and the \emph{mean} human rating $\bar{H}$ (a single series, not an
average of pairwise correlations). Used for the per-stratum and
score-region tables, where per-cell $n$ is too small to average three
pairwise correlations reliably.
\item \textbf{$\bar{H}$}: across-rater mean of the three human
ratings on a session $\times$ dimension.
\item \textbf{\% within 1}: proportion of sessions where the two
ratings being compared differ by at most 1 Likert point.
\item \textbf{recall (adversarial arm)}: proportion of perturbed
clips where the score drops by at least 1 point relative to the
same clip's baseline rating.
\item \textbf{L$>$H} / \textbf{H$>$L}: significance flag on a
(defect, dimension) cell where the LALM (respectively humans) are
more sensitive under a Newcombe-Wilson 95\% CI on the recall
difference.
\end{itemize}
